%% file: iclr2022_conference.tex
\documentclass{article} % For LaTeX2e
\usepackage{iclr2022_conference,times}

\input{math_commands.tex}

\usepackage{hyperref}
\usepackage{url}
\usepackage{booktabs}
\usepackage{subfig}
\usepackage{graphicx}
\usepackage{multirow}
\usepackage{enumitem}
\usepackage{float}

\newcommand{\HTLM}{HTLM}
\newcommand{\HTLiM  }{\texttt{CM3}} 

\makeatletter
    \newcommand{\vast}{\bBigg@{3}}
    \newcommand{\Vast}{\bBigg@{3.5}}
    \newcommand{\vastt}{\bBigg@{4}}
    \newcommand{\Vastt}{\bBigg@{4.5}}
\makeatother

\title{\HTLiM{}: A Causal Masked Multimodal Model of the Internet}

\iclrfinalcopy

\author{Armen Aghajanyan, Bernie Huang\thanks{Equal Contribution for Second Author} , Candace Ross\footnotemark[1] , Vlad Karpukhin\footnotemark[1] , Hu Xu\footnotemark[1], Naman Goyal,\\ \textbf{Dmytro Okhonko, Mandar Joshi, Gargi Ghosh, Mike Lewis, Luke Zettlemoyer} \\
Facebook AI Research\\
\texttt{\{armenag,berniehuang,ccross,vladk,huxu,naman\}@fb.com} \\
\texttt{\{oxo,mandarj,gghosh,mikelewis,lsz\}@fb.com} \\
}

\newcommand{\comment}[1]{}

\begin{document}

\maketitle

\begin{abstract}
We introduce \HTLiM{}, a family of causally masked generative models trained over a large corpus of structured multi-modal documents that can contain both text and image tokens. Our new causally masked approach  generates tokens left to right while also masking out a small number of long token spans that are generated at the end of the string, instead of their original positions. The casual masking object provides a type of hybrid of the more common causal and masked language models, by enabling full generative modeling while also providing bidirectional context when generating the masked spans.  
We train causally masked language-image models on large-scale web and Wikipedia articles, where each document contains all of the text, hypertext markup, hyperlinks, and image tokens (from a VQVAE-GAN), provided in the order they appear in the original HTML source (before masking). 
The resulting \HTLiM{} models can generate  rich structured, multi-modal outputs while conditioning on arbitrary masked document contexts, and thereby implicitly learn a wide range of text, image, and cross modal tasks. They can be prompted to recover, in a zero-shot fashion, the functionality of models such as DALL-E, GENRE, and \HTLM{} \citep{DALLE, GENRE, HTLM1}. We set the new state-of-the-art in zero-shot summarization, entity linking, and entity disambiguation while maintaining competitive performance in the fine-tuning setting. We can generate images unconditionally, conditioned on text (like DALL-E) and do captioning all in a zero-shot setting with a single model. 
\end{abstract}

\section{Introduction}
Recent advancements in large-scale generative sequence modeling have significantly improved zero-shot performance on multiple modalities, including text~\citet{gpt3,improving_few_shot_fb,HTLM1} and images~\citet{DALLE}.  Recent work has also shown how to use document structure, e.g., as provided by HTML web markup, to enable more effective zero-shot prompting for text-only tasks~\citep{HTLM1}. In this paper, we show it is possible to learn multi-modal document-structured generative models, to jointly represent formatted hypertext and images as they naturally co-occur within full document contexts. 

We introduce \HTLiM{}, a family of causally masked generative models trained over a large corpus of structured multi-modal documents. 
%To enable conditioning on arbitrary document structure, we introduce a new causally masked training objective. 
Causally masked models generate tokens left to right, just like a causal language model, but also mask out a small number of long token spans, which are then generated at the end of the string instead of their original positions. This provides a new hybrid of causal and masked language models, enabling full generative modeling with bidirectional context. For example, it can also be used in our setting to infill complete images or larger structured text sections, conditioned on the rest of the document.

We train \HTLiM{} models on close to a terabyte of web-based data following \citet{HTLM1}, extended to include images through VQVAE-GAN tokens \citep{vqvae_gan} and additional hypertext link structure. This data is in strong contrast to previous methods that were either uni-modal or carefully curated the image-text alignment (e.g., for image captioning~\citet{CLIP,DALLE}). We train a 2.7 billion and 13 billion causally masked model on this data which we call \HTLiM{}-Medium and \HTLiM{}-Large respectively.

Extensive experiments demonstrate that these models are able to perform a wide range of zero-shot uni- and cross-modal tasks. 
We show both qualitatively and quantitatively that \HTLiM{} can be prompted for non-trivial image generation, similar to that of DALL-E.
We also show that \HTLiM{} models are capable of improving over state-of-the-art zero-shot summarization, entity linking, entity disambiguation, highlighting the structure that comes from the hypertext during training. Finally, we show that by fine-tuning \HTLiM{} we set the new state-of-the-art for entity linking and entity disambiguation in general. 

To summarize, our contributions include:
\begin{itemize}
    \item We present the first hyper-text language-image model, trained on close to a Terabyte of multi-modal simplified HTML data from the common crawl. 
    \item We present the causally masked objective, a hybrid of causal and masked language models that allows for bidirectional context control during generative mask infilling.
    \item We demonstrate consistently strong transfer from \HTLiM{} to a range of uni-modal and multi-modal tasks at differing supervision levels, including stating state-of-the-art on entity disambiguation and zero-shot summarization.
    \item We release all code and models to support future \HTLiM{} research.
\end{itemize}

\section{Causally Masked Objective}
Traditional approaches to pre-training have focused on mixing the architectural choices (i.e., encoder-only, decoder-only, encoder-decoder) with objective choices (i.e., masking, causal language modeling). For example, masked encoder-only models such as BERT \citep{BERT} and RoBERTa \citep{ROBERTA} excel in non-generative fine-tuning tasks. Masked encoder-decoder models such as BART \citep{BART} and T5 \citep{T5} excel in both discriminative and generative fine-tuning. \citet{gpt3} on the other hand, showed that causal language models (de-facto, decoder only) are capable of non-trivial performance without the need of fine-tuning by simply prompting with appropriate string to control the generated outputs \cite{GPT2,gpt3,efficient_lm_xlm_g}. 

There are pros and cons to both masked and causal language modeling in the context of prompting. Masking offers the critical ability to encode bi-directionality within the prompts at the cost of only decoding roughly 15\% of the tokens of the input sequence during training \citep{BERT, ROBERTA, BART}. Conversely, decoder-only causal language models decode every token in the input sequence in the training but are typically limited to left-only contexts. Empirically, more work has also been done on scaling causal decoder-only rather than their counterparts.

In an effort to get most of the best of both worlds, we introduce a novel objective that combines the benefit of per-token generation with optional bi-directionality specifically tailored to prompting. For a document of size $s$ we select $n\sim \texttt{Clamp}(\texttt{Poisson(1)}, 1, 16)$ masks and for each of those masks we select span $m\sim(Uniform(0, s), Uniform(0, s))$ which does not intersect with any other $m$. These values are chosen to, on average, select relatively few relatively long spans, which we expect will allow the model to learn to infill long spans. We then order these masks by the order that they appear in the source document, replace the span of the mask in the source document with an enumerated mask token (i.e., \texttt{<mask:0>}, \texttt{<mask:1>}), and move the masked spans to the end of the document followed by a unique end of document token. 

\begin{figure}[h]
    \centering
    \includegraphics[width = 5.4in]{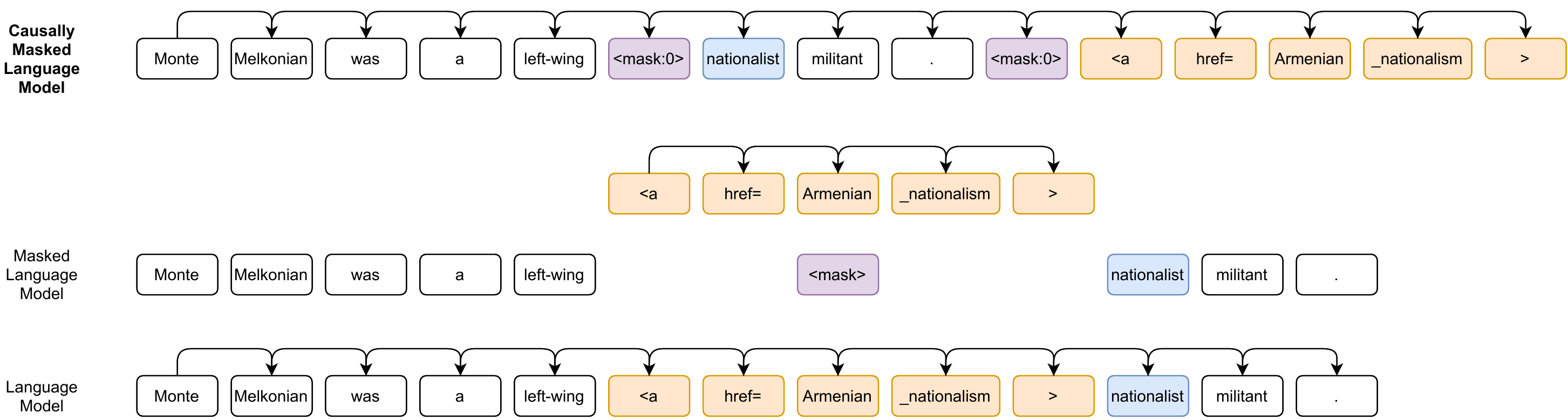}
    \caption{A visual representation of various language modeling objectives as well as our proposed causal language modeling objective with a single mask ($n=1$). Given the left-to-right nature of causal language models (bottom row) we would not be able to generate the Wikipedia entity link highlighted in orange.}
    \label{fig:types_of_masking}
\end{figure}

Figure~\ref{fig:types_of_masking} shows the complete process.

We also augment the standard cross-entropy loss to weigh the loss of predicting mask tokens as 0, as they are placed at random locations which carry no information to the underlying sequence modeling objective.

The complete array of benefits will become more apparent when designing prompts for uni/cross-modal tasks in \S~\ref{prompting}. However, at the core, the causally masked objective can do causal language modeling while optionally allowing for bidirectionality when needed. 
%The masking strategy here focuses on using a tiny number of masks with no restriction of size, which aligns better with the end task of prompting.

\section{\HTLiM{}}
\citet{HTLM1} used structured documents for text-only pre-training with strong zero-shot performance. \textbf{C}ausally-\textbf{M}asked \textbf{M}ultimodal \textbf{M}odeling (\HTLiM{}) extends this work by modeling full document structure including images and hypertext links. Furthermore, we move away from the BART-like objective of \citet{HTLM1} to use our new causally masked objective with decoder-only models.

\subsection{Data}
Following \citet{HTLM1} we aim to implement a transform over HTML documents to extract out to minimal-HTML, i.e., the minimal set of text that is semantically relevant for end tasks. 

\citet{laion_critic} gave in-depth criticisms of Common Crawl based multi-modal datasets and showed the existence of highly problematic examples (i.e., explicit images and text pairs of rape, pornography, and ethnic slurs). Given these severe ethical concerns, we opt-out of processing all of Common Crawl and instead opt into using a subset of the Common Crawl News (CC-NEWS) dataset and all of English Wikipedia. 

Given a valid HTML DOM\footnote{The DOM or Document Object Model is an interface that treats an HTML document as a tree structure wherein each node is an object representing a part of the document.} per document, we run several passes to strip down the DOM to the elements of maximal semantic value. We first remove all elements which do not contain textual elements.
We also filter out all \textit{headers}, \textit{footers}, \textit{copyrights}, \textit{forms}, \textit{dialog boxes} and \textit{iFrames}. We fold consecutive \verb+<div>+ elements into a singular \verb+<div>+ element with merged attributes. Furthermore we strip all the attributes from every element which are not derived from structured graphs such as OpenGraph, Schema and Twitter.

For every \texttt{<img>} tag in the document with a valid \texttt{src} attribute URL, we download the image, resize to 256x256 pixels with random cropping and then tokenize it with VQVAE-GAN from \citet{vqvae_gan}. This amounts to 256 tokens for every image. We then insert the string value of the tokens joined with a space back into the \texttt{src} attribute.

We do not place any restrictions on the number of images or their locations. We present a set of high-level statistics in Table~\ref{table:data_stats}.

\begin{table}[h]
\begin{tabular}{@{}lllll@{}}
\toprule
             & Documents (Million) & Size (GB) & Unique Images (Million) & Tokens (Billion) \\ \midrule
CC-NEWS      & 45                  & 460       & 18                      & 121              \\
En-Wikipedia & 16                  & 383       & 7                       & 102              \\ \midrule
Total        & 61                  & 843       & 25                      & 223 \\
\bottomrule
\end{tabular}
\caption{High level statistics of the data used to train \HTLiM{}.}
\label{table:data_stats}
\end{table}

For experimentation, we create two test sets from each data source with 10,000 unique documents for each. We de-duplicated our test sets to ensure no overlap between test and train sets to the best of our abilities.

\subsection{Size Hints}\label{sec:size_hints}
% hu: not sure if reader knows size hints well and then understand why not used in \HTLiM{}. If there's a load of reading/understanding maybe place this explanation in appendix?
\citet{HTLM1} introduced the concept of size hints which allows the user to guide the model during sample generation through token conditioning. Specifically, \HTLM{} inserts a probabilistic estimate of the size of the mask as a token post the mask token (e.g., \texttt{<mask>12} for a probabilistic size of 12). For \HTLiM{}, we noticed that size-hints degraded not only end-perplexity but also the zero-shot performance on a significant set of evaluation tests.

We also note that we can implicitly give a size hint during mask generation for a single mask by asking the model to generate causally \texttt{max\_sequence\_length - size\_hint} tokens before placing the secondary \texttt{<mask:0>} token.

\subsection{Training}
We train 4 models; 125M, 800M, 2.7B, and 13B parameters. The purpose of the two smaller models was to establish basic hyper-parameters that are viable for the causally masked language modeling objective and therefore were under-trained. \comment{They will be used in our perplexity analysis in \S~\ref{sec:ppl_analysis} of our multi-modal set-up.} However, all downstream tasks will be evaluated with our 2.7B model (\HTLiM{}-Medium) and our 13B model (\HTLiM{}-Large).
HTLM-Medium was trained on 240 V100 GPU for 28 days, while HTLM-Large was trained on 384 A100 GPU for 24 days. Our implementation was in PyTorch \citep{pytorch} using fairseq \citep{fairseq} and fairscale \citep{fairscale}. For every model, our per GPU batch size was 8, with a maximum token sequence length of 2048. We use the polynomial decay learning rate scheduler available in \citet{pytorch} with 1500 warmup updates. We clipped the gradient norms to 1.0 and used the Adam optimizer with $\beta_1=0.9$, $\beta_2=0.98$ \citep{ADAM}. We defer our model architecture description to \S~\ref{ssec:arch}.

\subsection{Scaling Laws}
Our training setting has a couple of new parameters that can impact the traditional scaling laws of causal language models. The multi-modal nature of our proposed model breaks the standard assumptions of token distributionality. Traditionally language tokens are said to follow a Zipfian distribution \citep{language_zipf}, while image tokens are strictly uniform (see \S~\ref{ssec:uniformity_tokens}). Furthermore, the unrestricted locations of the images and text introduce unpredictable complexity. Lastly, although we are still computing the joint probability of the document, we do so in a roundabout way through shuffling of the document via the causally masked objective. These fundamental differences warrant a quick look into the scaling laws of \HTLiM{}. 

\begin{figure}[h]
    \centering
    \includegraphics[width=\linewidth]{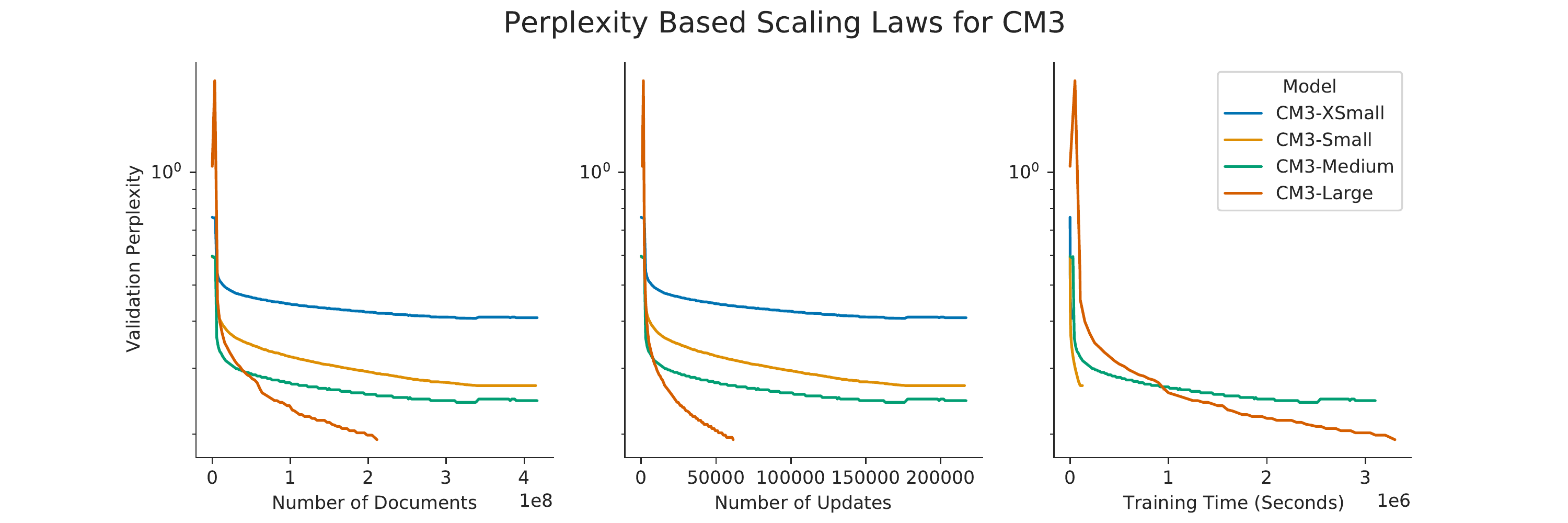}
    \caption{Basic perplexity based scaling laws for the proposed \HTLiM{} objective and training set-up.}
    \label{fig:scale_laws}
\end{figure}
We present the various perplexity curves for the four models of varying sizes we trained. Given that our models were trained on various hardware set-ups, we normalize the training time by linearly scaling each experiment timing to 256 GPU. Most importantly, we see healthy scaling, similar to \citet{scaling_laws_lm} without any pathological cases, implying there is still a good amount of gains to be achieved with further scaling. An in-depth analysis of the scaling laws of the causally masked objective is outside this current work's scope and will be considered for future work.

\section{Zero/Few-Shot Prompting}\label{prompting}
\subsection{Image Modality}
Although we do not train on pure image documents, \HTLiM{} can still operate over image tasks. To do so, we must cleverly describe the task through a textual prompt, using the \texttt{<img>} tag.
\subsubsection{Unconditional Image Generation}\label{sec:unconditional_image_generation}
To sample from the distribution of images available to \HTLiM{}, we can simply ask the model to produce the next set of tokens after the following prompt: \texttt{<img}.

Interestingly enough, \HTLiM{} prefers to first generate a short description of the image through the \texttt{alt} attribute and then generate the image tokens via the \texttt{src} attribute. We can force the model to directly generate image tokens without first giving a description with the following prompt: \texttt{<img src="}. We consider both prompts to test unconditional image generation since we do not condition the image generation but rather the model self-conditions. 

\begin{figure}[H]
\centering
\setlength{\tabcolsep}{3.0pt}
\begin{tabular}{lcccc}
\multirow{-9.4}{*}{\texttt{<img}} &
\subfloat[A mountain of olive trees on the way to Cabo de la Vela]{\includegraphics[width = 0.95in]{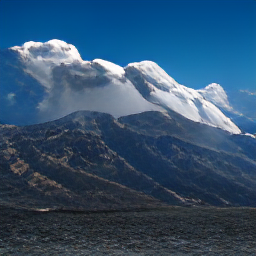}} &
\subfloat[Spain Europa Amenacer Winter]{\includegraphics[width = 0.95in]{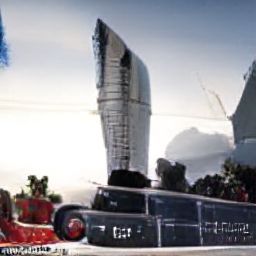}} &
\subfloat[blog TIGI Bed Head Tie Dye Spray Hair Spray Hairspray ml]{\includegraphics[width = 0.95in]{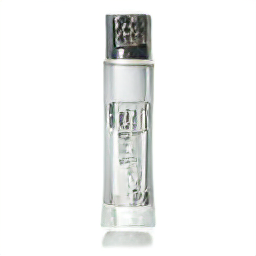}} &
\subfloat[birthday invitation printable christmas gift for birthday party Printable Template]{\includegraphics[width = 0.95in]{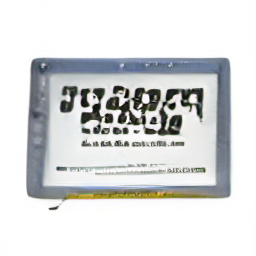}} \\ 

\multirow{-9.4}{*}{\texttt{<img src="}} &
\subfloat{\includegraphics[width = 0.95in]{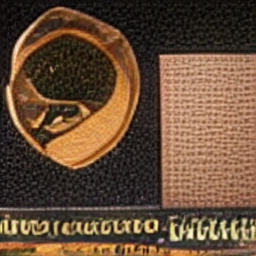}} &
\subfloat{\includegraphics[width = 0.95in]{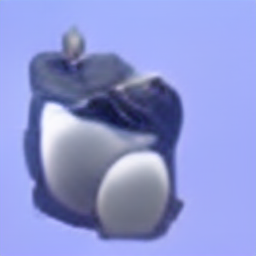}} &
\subfloat{\includegraphics[width = 0.95in]{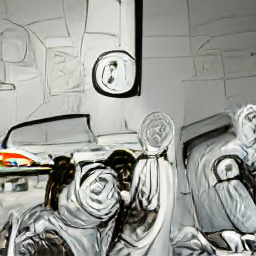}} &
\subfloat{\includegraphics[width = 0.95in]{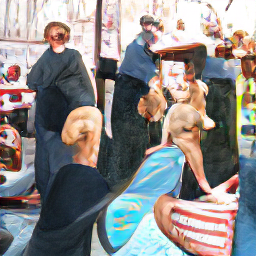}}

\end{tabular}
\caption{Four samples for two of the prompts we proposed for unconditional image generation for \HTLiM{}-Large. For the self-captioned images we place the respective caption under the image. Results were selected at random, with no cherry picking.}
\label{fig:unconditional_image_generation}
\end{figure}

We sample according to the distribution of the model without altering the temperature. We present a sample of non-cherry picked examples in Figure~\ref{fig:unconditional_image_generation}. 

The model is more than capable of generating coherent images. We note that via this prompting, we can recover the full functionality of the DALL-E model proposed in \citet{DALLE}. Interestingly enough, we see qualitative improvements with allowing the model to free generate a caption prior to generating. 

We continue by doing an empirical study of the unconditional generation of \HTLiM{}, by generating 30k samples without textual conditioning and calculating the Fréchet Inception Distance (FID, \citet{two_time_gan}) over MS-COCO, following the methodology proposed in \citet{ GLIDE} \citep{mscoco}. We present our results in the unified table showing FID calculations in Table~\ref{table:mscoco_fids}. Without any textual conditioning and without explicitly optimizing for either MS-COCO or generation (unlike other benchmarks in the table) \HTLiM{} Large approaches the FID performance of modern Generative Adversarial Networks (GAN).

\subsubsection{Image In-Filling}
Unlike DALL-E, which leverages left-to-right language modeling objective to model language-image tokens, 
\HTLiM{} with the proposed causally masked language modeling makes it possible to condition contiguous sections of an image on the surrounding context for image in-filling. Specifically, \HTLiM{} can infill images with the following prompt:

\begin{center}
\small
    \begin{quote}
        \texttt{\textbf{Infilling Prompt:} <img src="\{prefix\}<mask:0>\{postfix\}"><mask:0>}
    \end{quote}
\end{center}

Using the same decoding strategies described in \S~\ref{sec:unconditional_image_generation} we generate unconditional infilled images with only \HTLiM{}-Large and present qualitative results in Figure~\ref{fig:qualitative_infilling}. Overall we see that \HTLiM{}-Large is capable of generating semantically coherent infills even without grounding in text.

\subsection{Text-Image Modality}
\subsubsection{Conditional Image In-Filling}
Additionally, \HTLiM{} can further perform image in-filling condition on the additional text context. This can be achieved by slightly augmenting the prompt as follows:
\begin{center}
\small
    \begin{quote}
        \texttt{\textbf{Conditional Infilling Prompt:}}
    \end{quote}
    \begin{quote}
        \texttt{<img alt="Photo: \{text\}" src="\{prefix\}<mask:0>\{postfix\}"><mask:0>}
    \end{quote}
\end{center}

\begin{figure}[ht]
\centering
\small
\setlength{\tabcolsep}{4.0pt}
\begin{tabular}{ccccr}
\subfloat{\includegraphics[width = 0.95in, height = 0.95in]{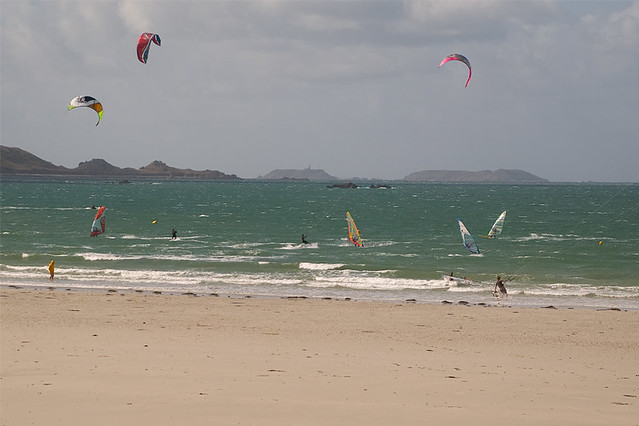}} &
\subfloat{\includegraphics[width = 0.95in, height = 0.95in]{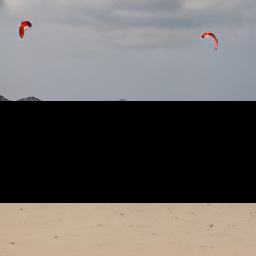}} &
\subfloat{\includegraphics[width = 0.95in, height = 0.95in]{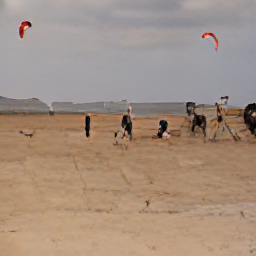}} &
\subfloat{\includegraphics[width = 0.95in, height = 0.95in]{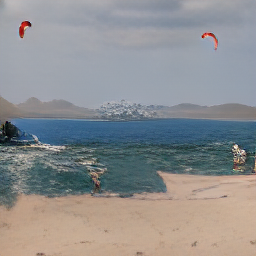}}
& \multirow{-9.4}{2.5cm}{\scriptsize \texttt{group of people windsurfing over the beach and water in the ocean.}} \\

\subfloat{\includegraphics[width = 0.95in, height = 0.95in]{resources/infilling/infilling_1_0.jpeg}} &
\subfloat{\includegraphics[width = 0.95in, height = 0.95in]{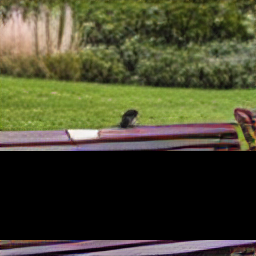}} &
\subfloat{\includegraphics[width = 0.95in, height = 0.95in]{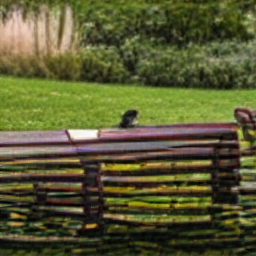}} &
\subfloat{\includegraphics[width = 0.95in, height = 0.95in]{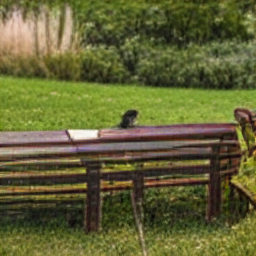}} & \multirow{-9.4}{2.5cm}{\scriptsize \texttt{the wooden park benches are painted dark purple.}} \\

\subfloat{\includegraphics[width = 0.95in, height = 0.95in]{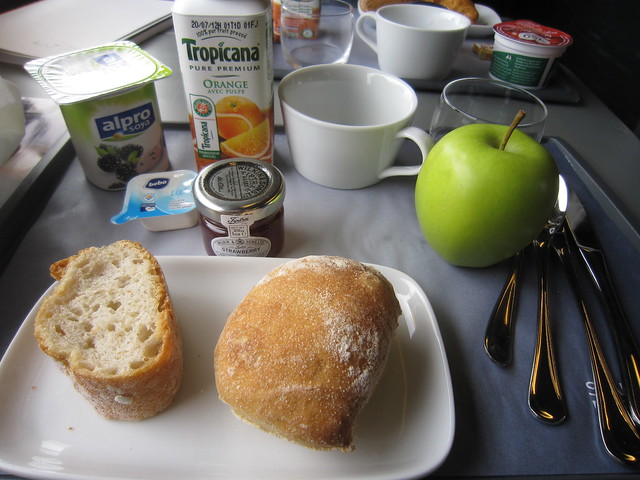}} &
\subfloat{\includegraphics[width = 0.95in, height = 0.95in]{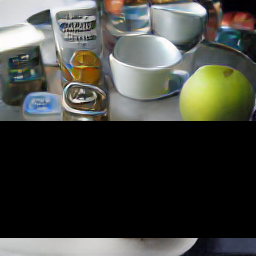}} &
\subfloat{\includegraphics[width = 0.95in, height = 0.95in]{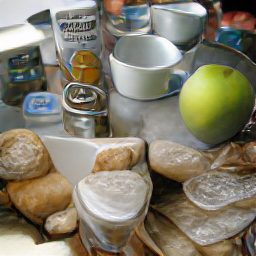}} &
\subfloat{\includegraphics[width = 0.95in, height = 0.95in]{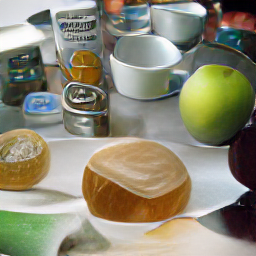}} & 
\multirow{-9.4}{2.5cm}{\scriptsize \texttt{some bread is on a plate with jam, an apple, yogurt and orange juice.}} \\

\subfloat{\includegraphics[width = 0.95in, height = 0.95in]{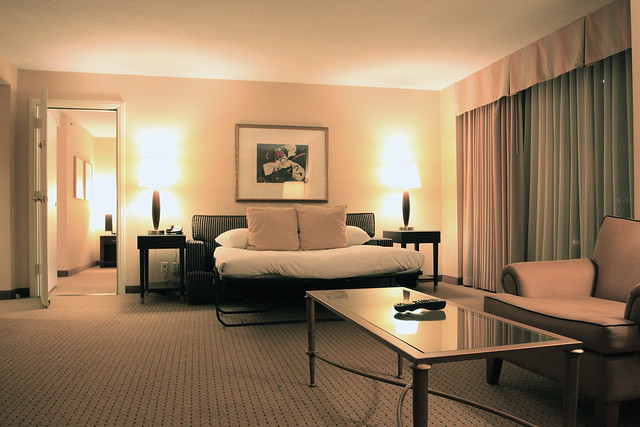}} &
\subfloat{\includegraphics[width = 0.95in, height = 0.95in]{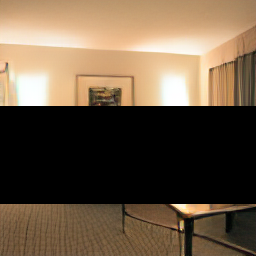}} &
\subfloat{\includegraphics[width = 0.95in, height = 0.95in]{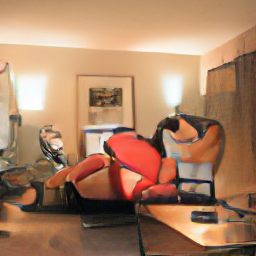}} &
\subfloat{\includegraphics[width = 0.95in, height = 0.95in]{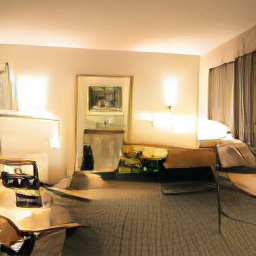}} & 
\multirow{-9.4}{2.5cm}{\scriptsize \texttt{a nice looking hotel room with a neatly done bed, coffee table, and a chair.}} \\ \midrule
\multicolumn{1}{c}{Source Image}  & \multicolumn{1}{c}{Masked/Tokenized Image} & \multicolumn{1}{c}{\HTLiM{}-Infilling-U} & \multicolumn{1}{c}{\HTLiM{}-Infilling-C} & \multicolumn{1}{c}{Ground Truth}

\end{tabular}
\caption{We provide qualitative samples for zero-shot image-infilling using the \HTLiM{}-Large model using the aforementioned prompts. \HTLiM{}-Infilling-U refers to infilling without conditioning on text while \HTLiM{}-Infilling-C refers to conditioning on the ground truth text.}
\label{fig:qualitative_infilling}
\end{figure}

We show qualitative results in Figure~\ref{fig:qualitative_infilling}. Immediately we notice the substantial improvement in the generated image when grounded in ground truth text vs. unconditional image-infilling.

\subsubsection{Conditional Image Generation}\label{sec:conditional_image_generation}
We can do conditional text generation using \HTLiM{} similar to DALL-E by using a proper prompt. Specifically by conditioning using the \texttt{alt} attribute of the \texttt{img} tag.
\begin{center}
\small
    \begin{quote}
        \texttt{\textbf{Conditional Generation Prompt:} <img alt="\{prompt\}}
    \end{quote}
\end{center}

\begin{figure}[h]
\setlength{\tabcolsep}{3.0pt}
\centering
\begin{tabular}{lrrrr}
\multirow{-9.4}{2.5cm}{\small \centering \texttt{an armchair in the shape of an avocado. an armchair imitating an avocado.}} &
\subfloat{\includegraphics[width = 0.95in]{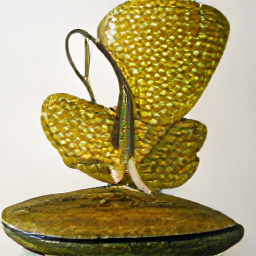}} &
\subfloat{\includegraphics[width = 0.95in]{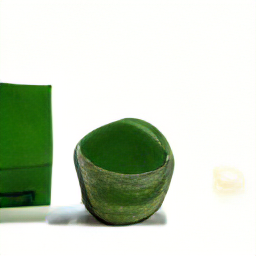}} &
\subfloat{\includegraphics[width = 0.95in]{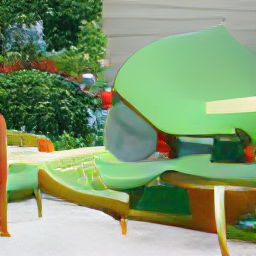}} &
\subfloat{\includegraphics[width = 0.95in]{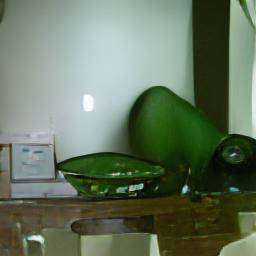}} \\

\multirow{-9.4}{2.5cm}{\small \centering \texttt{A red car in the mountains.}} &
\subfloat{\includegraphics[width = 0.95in]{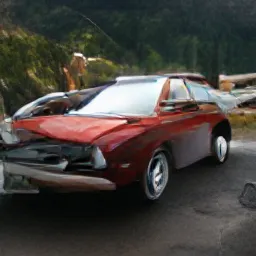}} &
\subfloat{\includegraphics[width = 0.95in]{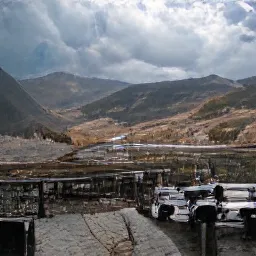}} &
\subfloat{\includegraphics[width = 0.95in]{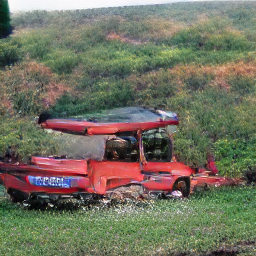}} &
\subfloat{\includegraphics[width = 0.95in]{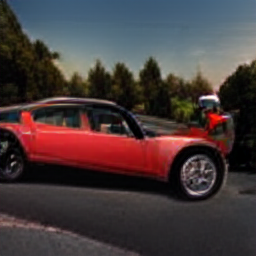}} \\

\multirow{-9.4}{2.5cm}{\small \centering \texttt{Photo: A sheep in snowy Artsakh}} &
\subfloat{\includegraphics[width = 0.95in]{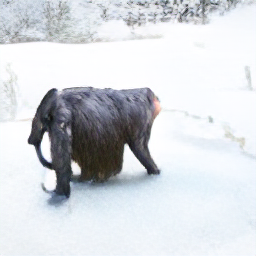}} &
\subfloat{\includegraphics[width = 0.95in]{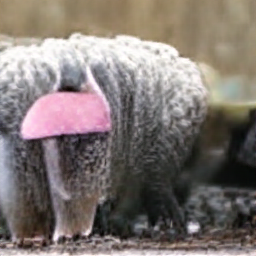}} &
\subfloat{\includegraphics[width = 0.95in]{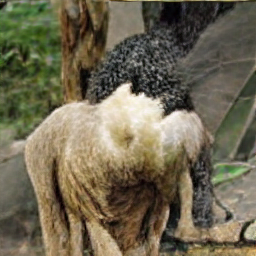}} &
\subfloat{\includegraphics[width = 0.95in]{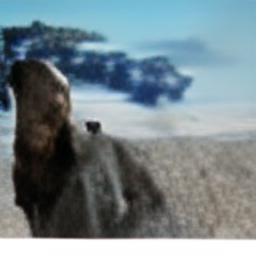}} \\

\multirow{-9.4}{2.5cm}{\small \centering \texttt{Photo: An Armenian church during springtime with clear skies}} &
\subfloat{\includegraphics[width = 0.95in]{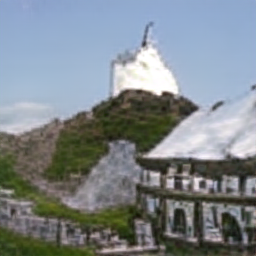}} &
\subfloat{\includegraphics[width = 0.95in]{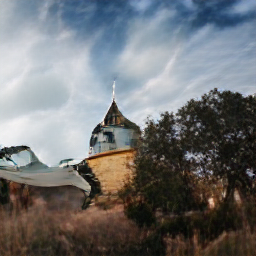}} &
\subfloat{\includegraphics[width = 0.95in]{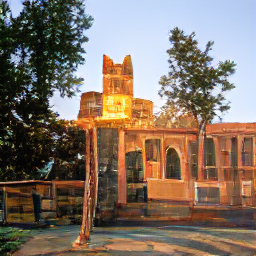}} &
\subfloat{\includegraphics[width = 0.95in]{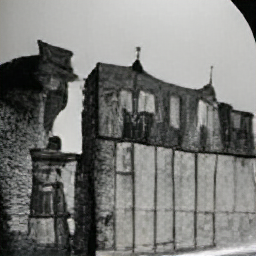}}

\end{tabular}
\caption{Four samples for four of the prompts using the conditional image generation prompt with \HTLiM{}-Large. Results were selected by CLIP from a candidate set of 32 samples.}
\label{fig:conditional_image_generation}
\end{figure}

We present qualitative conditional image generation results in Figure~\ref{fig:conditional_image_generation}. Specifically, we sample 32 images for every prompt given and re-rank using CLIP to get the top-4 images \citep{CLIP}. Overall we see that \HTLiM{} can generate recognizable images of the input text. There are still failure cases, such as the second image in the second prompt, where the model easily generates a landscape but forgets to generate the red car. The third prompt, \HTLiM{}, is incapable of drawing the face of a sheep while getting the general body and texture correct.

We note that \HTLiM{} trains with an order of magnitude less unique images than DALL-E, and the subset of images available to \HTLiM{} are the images available in news and Wikipedia articles; therefore, \HTLiM{} does not generate fictional images well. That being said, casting a larger pool for CLIP selection by randomly sampling a larger set qualitatively fixes some of these issues. 

\begin{table}[t]
    \centering
    \begin{center}
    \begin{small}
    \begin{tabular}{lrr}
    \toprule
    Model & FID & Zero-shot FID \\
    \midrule
    AttnGAN \citep{attngan} & 35.49 & \\
    DM-GAN \citep{dmgan} & 32.64 & \\
    DF-GAN \citep{dfgan} & 21.42 & \\
    DM-GAN + CL \citep{textcl} & 20.79 & \\
    XMC-GAN \citep{xmcgan} & 9.33 & \\
    LAFITE \citep{lafite} & \textbf{8.12} & \\
    DALL-E \citep{DALLE} & & $\sim$ 28 \\
    LAFITE \citep{lafite} & & 26.94 \\
    GLIDE \citep{GLIDE} & & \textbf{12.24} \\
    \midrule
    Unconditional \HTLiM{}-Medium &  & 40.65 \\
    Unconditional \HTLiM{}-Medium &  & 36.51 \\
    Conditional \HTLiM{}-Medium &  & 36.78 \\
    Conditional \HTLiM{}-Large & & 29.56 \\
    \bottomrule
    \end{tabular}
    \end{small}
    \end{center}
    \caption{We compare FID on MS-COCO $256 \times 256$. Following \citet{GLIDE} we sample roughly 30k conditioned samples for our models, and compare against the entire validation set. We use a temperature of 0.85 for both \HTLiM{} models. We use the implementation available from \citet{FID_implementation}.}
    \label{table:mscoco_fids}
\end{table}

For quantitative analysis, we compute FID on MS-COCO following the methodology provided by \citet{GLIDE}. Specifically, we sample 30k samples conditioned on MS-COCO captions. For all models, we use a temperature of 0.85 and do straightforward sampling.

We present our FID results on MS-COCO 256x256 in Table~\ref{table:mscoco_fids}. In general \HTLiM{} is capable of generating semantically coherent images on-par with modern GANs. Furthermore, our conditional \HTLiM{}-Large model approaches the performance of the DALL-E model while using an order of magnitude fewer data.
\subsubsection{Captioning}
We next look at the dual-task to conditional image generation and image captioning. We can prompt \HTLiM{} to do zero-shot image captioning by asking the model to generate either the \texttt{alt} or \texttt{title} attributes of a properly set \texttt{<img>} tag. Due to attributes always appearing in alphabetical order in order to generate \texttt{alt} attribute (which appears before \texttt{src}), we need to use the masking capabilities of \HTLiM{}.

\begin{center}
\footnotesize
    \begin{quote}
        \texttt{\textbf{Captioning Masked Prompt \#1:} <img alt="Photo: A photo taken of<mask:0>" src="\{image\}">}
    \end{quote}
\end{center}

\begin{center}
\small
    \begin{quote}
        \texttt{\textbf{Captioning Causal Prompt \#1:} <img src="\{image\}" title="Photo: A photo taken of}
    \end{quote}
\end{center}

We have two methods of generating captions given the above prompts. First, the relatively inexpensive method involves running beam-search with a beam size of 5 over the proposed prompts. For a single image, we run both available prompts and select the sequence, which minimizes the respective \HTLiM{} perplexity. The second method is much more costly and requires sampling 128 captions for every image (we note that this is cheaper than image generation since image generation requires the minimal generation of 256 image tokens while captioning is usually on the order of a dozen tokens). We then use CLIP from \citet{CLIP} to get the top ranking caption. We note that non-trivial captioning behavior was only exhibited in \HTLiM{}-Large model; therefore, all evaluations will consider this singular model.

\begin{figure}[H]
\centering
\small
\setlength{\tabcolsep}{4.0pt}
\begin{tabular}{rrrrr}
\subfloat{\includegraphics[width = 0.95in, height = 0.95in]{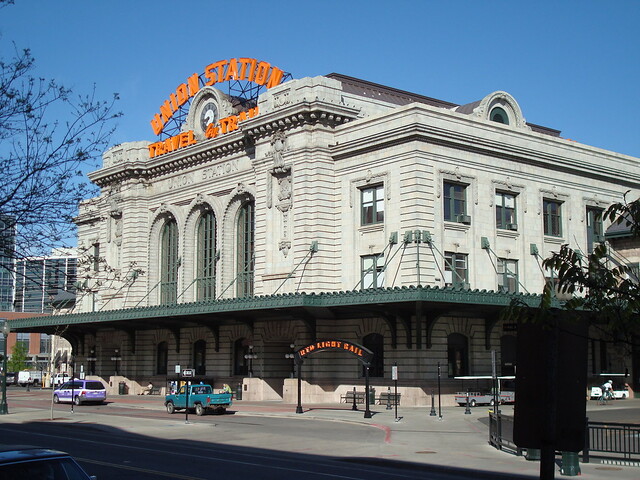}} &
\subfloat{\includegraphics[width = 0.95in, height = 0.95in]{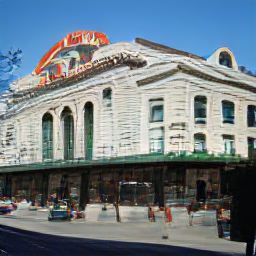}} &
\multirow{-9.4}{2.5cm}{\scriptsize \texttt{the main entrance of the U.S. Department of State in Washington, D.C.}} &
\multirow{-9.4}{2.5cm}{\scriptsize \texttt{the white marble exterior standing atop of its façade.}} 
& \multirow{-9.4}{2.5cm}{\scriptsize \texttt{outside of a train station building from across the street.}} \\

\subfloat{\includegraphics[width = 0.95in, height = 0.95in]{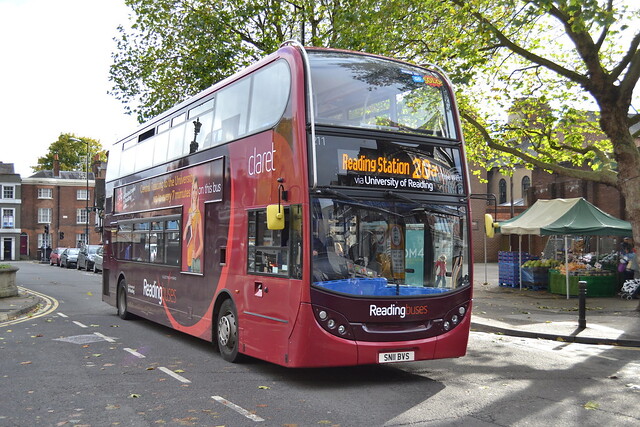}} &
\subfloat{\includegraphics[width = 0.95in, height = 0.95in]{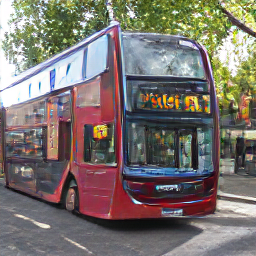}} &
\multirow{-9.4}{2.5cm}{\scriptsize \texttt{a pickup truck parked in a layby on a highway.}} &
\multirow{-9.4}{2.5cm}{\scriptsize \texttt{a large bus parked in a layby}} & \multirow{-9.4}{2.5cm}{\scriptsize \texttt{a tall red bus is coming down some tracks}} \\

\subfloat{\includegraphics[width = 0.95in, height = 0.95in]{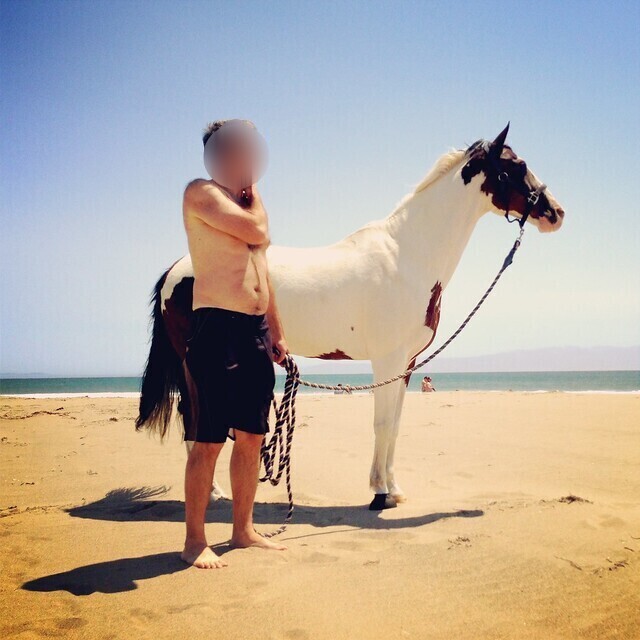}} &
\subfloat{\includegraphics[width = 0.95in, height = 0.95in]{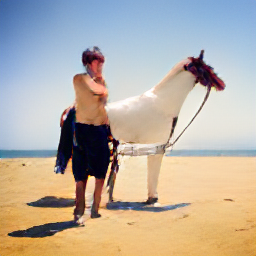}} &
\multirow{-9.4}{2.5cm}{\scriptsize \texttt{a man posing for a photo.}} &
\multirow{-9.4}{2.5cm}{\scriptsize \texttt{a man next to a large horse.}} & 
\multirow{-9.4}{2.5cm}{\scriptsize \texttt{a man standing next to a horse on a beach
}} \\

\subfloat{\includegraphics[width = 0.95in, height = 0.95in]{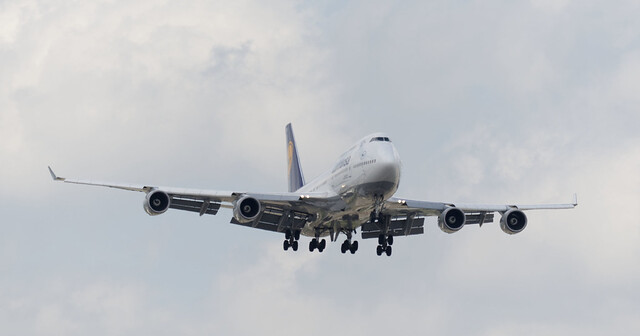}} &
\subfloat{\includegraphics[width = 0.95in, height = 0.95in]{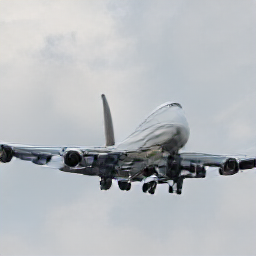}} &
\multirow{-9.4}{2.5cm}{\scriptsize \texttt{a U.S. Air Force B-52H Stratofortress on the flight line at Barksdale Air Force Base, Louisiana}} &
\multirow{-9.4}{2.5cm}{\scriptsize \texttt{the Austrian Airbus A321 aircraft with its Austrian registration}} & 
\multirow{-9.4}{2.5cm}{\scriptsize \texttt{a jet airliner flying with a cloudy sky in the background.}} \\ \midrule
\multicolumn{1}{c}{Source Image}  & \multicolumn{1}{c}{Tokenized Image} & \multicolumn{1}{c}{\HTLiM{}-Caption-Beam} & \multicolumn{1}{c}{\HTLiM{}-Caption-CLIP} & \multicolumn{1}{c}{Ground Truth}

\end{tabular}
\caption{We provide qualitative samples for zero-shot image-captioning using the \HTLiM{}-Large model. Caption-Beam refers to generating caption using beam over prompts, while Caption-CLIP uses CLIP to get the top-ranked caption from a 128 candidate set (64 from masked prompt, 64 from causal prompt).}
\label{fig:qualitative_captioning}
\end{figure}

We provide a qualitative example in Figure~\ref{fig:qualitative_captioning}, sourcing images and ground truth captions from MS-COCO \citep{mscoco}. We see that \HTLiM{} is capable of generating non-trivial semantically coherent captions. That being said, most failure cases of our proposed zero-shot captioning are due to the loss of texture from representing images through discrete tokens (e.g., the text of the train station is blurred, as is the text on the bus).

%  title="Photo Description: A photo taken of 
% Photo: A photo taken 

Quantitatively we measure the quality of \HTLiM{} zero-shot captioning by evaluating using BERT-Score\footnote{We use the open-source BERTScore at: \url{https://github.com/Tiiiger/bert_score}. The evaluation method is: \texttt{roberta-large\_L17\_no-idf\_version=0.3.11(hug\_trans=4.11.3)\_fast-tokenizer}} \citep{bertscore} with the RoBERTa-Large models \citep{ROBERTA} on the validation set from MS-COCO. We opt for the use of semantic evaluation versus classical metrics such as BLEU/METEOR because we notice that the vocabulary and sentence structure 
% the style is not compatible
of zero-shot captioning with \HTLiM{} is not compatible with MS-COCO ground truth labels, although the generated content is semantically similar.
We present our quantitative result in Table~\ref{table:caption}. \HTLiM{}-Large is capable of achieving reasonable zero-shot captioning performance on the MS-COCO dataset.
% Consider adding other baselines in table 3.

\begin{table}[h]
\centering
\begin{tabular}{@{}llll@{}}
\toprule
                & Precision & Recall & F1    \\ \midrule
\HTLiM{}-Caption-Beam & 0.781     & 0.789  & 0.785 \\
\HTLiM{}-Caption-CLIP & 0.863     & 0.866  & 0.864 \\ \bottomrule
\end{tabular}
\caption{BERTScore numbers for zero-shot captioning with \HTLiM{}.}\label{table:caption} 
\end{table}

\subsection{Text Modality}
\HTLiM{} is not only a cross-modal model but is fully capable of acting as a stand-alone language model. This is even reflected in our data, where we do not enforce every document to have images; therefore, pure language modeling will also occur during training. We evaluate our \HTLiM{} models on a wide set of varying language tasks.

\subsubsection{Entity Disambiguation}
We reproduce the evaluation setting described by \citet{GENRE} and \citet{improving_entity_disambiguation} using the same candidate sets, datasets and evaluating using the InKB micro-F1 metric. 

We aim to find a prompt capable of representing the more general end-to-end entity linking task in the \HTLiM{} model. From there, a proper sequence scoring of the candidate set will provide us with an approach to zero-shot entity disambiguation. Luckily HTML based Wikipedia contains very rich annotations. Specifically below, we show an example of naturally occurring entity linking that would occur in our Wikipedia subset of \HTLiM{} training data.

\begin{quote}
    \texttt{\textbf{Original:} \textit{Manetho} writes that these kings ruled from <a title="Memphis, Egypt">\textit{Memphis}</a>}
\end{quote}
\begin{quote}
    \texttt{\textbf{Prompt:} \textit{Manetho} writes that these kings ruled from <a title="<mask:0>">\textit{Memphis}</a>...<mask:0>}
\end{quote}
\begin{quote}
    \texttt{\textbf{Target:} \textit{Manetho} writes that these kings ruled from <a title="<mask:0>">\textit{Memphis}</a>...<mask:0> Memphis, Egypt}
\end{quote}

Using our scoring approach we can simply score the \textbf{Target} while swapping out the postfix after \texttt{<mask:0>}.

\begin{table}[h]
\centering
\resizebox{\textwidth}{!}{   

\fontsize{8.4}{10.1}\selectfont \setlength{\tabcolsep}{0.5em}
\begin{tabular}{llcccccc|c}
\toprule
& & In-domain  & \multicolumn{5}{c}{Out-of-domain} &  \\
\textbf{Model Type} & \textbf{Method} &  \textbf{AIDA} &  \textbf{MSNBC} &  \textbf{AQUAINT} &  \textbf{ACE2004} &  \textbf{CWEB} &  \textbf{WIKI*} &  \textbf{Avg.} \\
\midrule
\multirow{8}{*}{\emph{Direct Supervision} \quad \quad \ \Vastt\{ } & \citet{ganea2017deep}  & 92.2 & 93.7 & 88.5 & 88.5 & 77.9 & 77.5 & 86.4 \\
& \citet{robust_entity_disambiguation_random_walk}  & 89 & 92 &   87 &   88 & 77 & {84.5} & 86.2 \\
& \citet{yang2018collective} & \textbf{95.9} & 92.6 &  89.9 & 88.5 & \textbf{81.8} & 79.2 & {88.0} \\
& \citet{shahbazi2019entity}        & 93.5 & 92.3 &   {90.1} &  88.7 & {78.4} & 79.8 & 87.1 \\
& \citet{yang2019learning}   & 93.7 & {93.8} & 88.2 & {90.1} & 75.6 & 78.8 & 86.7 \\
& \citet{le2019boosting} & 89.6 & 92.2 & 90.7 &  88.1 & 78.2 & 81.7 & 86.8 \\
& \citet{fang2019joint}  & {94.3} & 92.8 &  87.5 & 91.2 & 78.5 & 82.8 & 87.9 \\
& \textbf{\cite{GENRE}}      &      93.3 & 94.3 & 89.9 & {90.1} & 77.3 & \textbf{87.4} & 88.8 \\ \midrule
\multirow{2}{*}{\emph{Direct Supervision} \quad \quad \  \  \Large\{ }  & \HTLiM{}-Medium     &      93.5 & 94.2 & 90.1 & 90.4 & 76.5 & 86.9 & 88.6 \\
& \HTLiM{}-Large      &      94.8 & \textbf{94.8} & \textbf{91.1} & \textbf{91.4} & 78.4 & \textbf{88.7} & \textbf{89.8} \\ \midrule
\multirow{2}{*}{\emph{Self Supervision (0-Shot)} \Large\{ } & \HTLiM{}-Medium      &      78.0 & 80.1 & 75.4 & 81.4 & 68.5 & 76.2 & 76.6 \\
& \HTLiM{}-Large      &      80.1 & 80.8 & 77.7 & 82.8 & 72.4 & 80.2 & 79.0 \\ \bottomrule
\end{tabular} 
}

\caption{Aligned with GENRE's evaluation, we use Micro $F_1$ (InKB) for the named entity disambiguation task. \textbf{Bold} indicates best model. We note that although *WIKI can be thought of as being out-of-domain, given that English Wikipedia was used to pre-train \HTLiM{}, it can be considered in-domain as well.
}
\label{table:ned-results}
\end{table}

As an additional datapoint for the representations learned from \HTLiM{} we completely replicate the training and evaluation for the GENRE model \citep{GENRE}.\footnote{\url{https://github.com/facebookresearch/GENRE}}. Specifically we first fine-tune \HTLiM{} on the BLINK data \citep{zero_shot_entity_linking}. For the in-domain scenario, we fine-tune \HTLiM{} on the AIDA-CoNLL dataset \citep{robust_entity_disambiguation}. We evaluate on the AIDA-CoNLL dataset for the in-domain scenario and the MSNBC, AQUAINT,
ACE2004, WNED-CWEB (CWEB) and WNED-WIKI (WIKI) for the out-of-domain scenario \citep{GENRE, robust_entity_disambiguation_random_walk}. We present our results in Figure~\ref{table:ned-results}.

Given the strong supervision naturally available in Wikipedia HTML, it is unsurprising that \HTLiM{} shows strong, non-trivial zero-shot performance on the named entity disambiguation across a wide array of named entity disambiguation tasks. 

Furthermore, the fine-tuned HTLM-Large model outperforms previous entity linking specific models to achieve a new SOTA over the benchmarked datasets.

\subsubsection{Entity Linking}
We next consider the more general entity linking task. We experiment with two settings zero-shot assuming we know the location of the entities and the full fine-tuning setting following the exact methodology proposed in \citet{GENRE}. Specifically for the end-to-end Entity Linking, we aim to reproduce the setting of~\citet{kolitsas2018end}. We evaluate using the aforementioned  \textit{InKB} micro-$F_1$ with the same defined in-domain and out-of-domain datasets as described by \citet{GENRE}. We use the exact same \emph{in-domain} and \emph{out-of-domain} datasets as well as evaluating the \textit{InKB} micro-$F_1$ on the GERBIL benchmark platform~\citep{roder2018gerbil}. Furthermore, we use the same decoding strategy for the zero-shot case by limiting the generative tokens to only available candidate entities. Please refer to \citet{GENRE} for the full fine-tuning setup.

For both setting we evaluate on seven test sets: MSNBC, Derczynski (Der)~\citep{derczynski2015analysis}, KORE 50 (K50)~\citep{hoffart2012kore}, N3-Reuters-128 (R128), N3-RSS-500 (R500)~\citep{roder2014n3}, and OKE challenge 2015 and 2016 (OKE15 and OKE16)~\citep{nuzzolese2015open}.

\begin{table}[h]
\centering
\resizebox{\textwidth}{!}{
\fontsize{8.4}{10.1}\selectfont \setlength{\tabcolsep}{0.5em}
\begin{tabular}{llcccccccc|c}
\toprule
& & In-domain  & \multicolumn{7}{c}{Out-of-domain}  \\
 & \textbf{Method} & \textbf{AIDA} & \textbf{MSNBC} &  \textbf{Der} &  \textbf{K50} & \textbf{R128} & \textbf{R500} & \textbf{OKE15*} & \textbf{OKE16*} & \textbf{Avg.}\\
\midrule
\multirow{8}{*}{\emph{Direct Supervision} \quad \quad \ \Vastt\{ } & \citet{robust_entity_disambiguation}     &   72.8 &   65.1 &       32.6 &   55.4 &          46.4 &       \textbf{42.4} &     \textbf{63.1} &      0.0 & 47.2 \\
& \citet{steinmetz2013semantic} &   42.3 &   30.9 &       26.5 &  46.8  &   18.1 &       20.5 &     46.2 &     46.4 & 34.7 \\
& \citet{moro2014entity}        &   48.5 &   39.7 &       29.8 &   55.9 &  23.0 &       29.1 &  41.9 &     37.7 & 38.2 \\
& \citet{kolitsas2018end}       &   82.4 &   72.4 &       34.1 &   35.2&  \textbf{50.3} &       38.2 &     61.9 &     52.7 & 53.4\\
& \citet{broscheit2020investigating} &   79.3 &    - &        - &    - & - &    - &  - &    -  \\
& \citet{martins2019joint}      &   81.9 &    - &        - & - &    -&    - &    - &  - \\
& \citet{van2020rel}$^\dagger$      &   80.5 &   72.4 &       41.1 &   50.7 & 49.9 & 35.0  & \textbf{63.1} & \textbf{58.3} & 56.4 \\
& \citet{GENRE} &   \textbf{83.7} &   73.7 & \textbf{54.1} &   60.7 & 46.7 & 40.3 & 56.1 & 50.0 & \textbf{58.2} \\ \midrule \multirow{2}{*}{\emph{Direct Supervision} \quad \quad \ \ \Large\{ } & 
\HTLiM{}-Medium     &   71.4 &   68.5 &       48.6 &   58.3 &          44.9 &       41.1 &     61.9 &      37.7 & 54.1 \\
& 
\HTLiM{}-Large     &   79.9 &   \textbf{74.8} &       53.2 &   \textbf{62.4} &          47.1 &       \textbf{42.8} &     61.9 &      52.7 & \textbf{59.3} \\
\midrule \multirow{2}{*}{\emph{Self Supervision (0-Shot)} \Large\{ } & 
\HTLiM{}-Medium     &   20.4 &   18.6 &       20.1 &   35.1 &          30.6 &       32.1 &     36.6 &      0.0 & 24.2 \\
& 
\HTLiM{}-Large     &   24.8 &   21.4 &       25.6 &   39.0 &          31.1 &       34.9 &     37.1 &      0.0 & 26.7 \\
\bottomrule

\end{tabular}
}
\caption{We report Micro $F_1$ on our test sets for our entity linking task. \textbf{Bold} indicates best model. Following \citet{GENRE}
we use a $^\dagger$ to indicate results from the Wikipedia 2019 setting as opposed to the 2014 setting (which has older dump and fewer entities).}
\label{table:el-results}
\end{table}

We present our results in Table~\ref{table:el-results}. We see that our \HTLiM{} are extremely competitive with entity-linking specific models and that our \HTLiM{}-Large model sets a new state-of-the-art. Furthermore, although our zero-shot numbers are substantially worse, they are still non-trivial, implying that \HTLiM{} learns a significant amount of implicit entity linking through our training setting.

\subsubsection{Summarization}
We next look at \HTLiM{} performance on the zero-shot summarization task, specifically we replicate the zero-shot evaluation methodology of \citet{HTLM1}. For all summarization benchmarks, we use ROUGE-1/2/L as our primary metrics to stay consistent with other literature~\citep{ROUGE}. We look at the same datasets as \HTLM{}.

\begin{itemize}[label={},leftmargin=0pt]
    \item \textbf{Gigaword} consists of headlines from news articles~\citep{gigaword}. The target summaries are relatively short, consisting roughly on average of 10 BPE tokens.

    \item \textbf{CNN/Dailymail}~\citep{cnndailymail}
 provides multi-sentence target summaries close to 3 sentences, or roughly 50 tokens.

    \item \textbf{Reddit TIFU}~\citep{reddittifu}
contains summaries of Reddit posts. Specifically, we use the \textit{short} subset of data . Compared to our other summarization datasets, this dataset is highly abstractive and not based on news articles.

    \item \textbf{XSum} \citep{xsum} provides abstractive single sentence summaries of news articles.
\end{itemize}

We utilize the same prompts as available in \citet{HTLM1}. We use the same available size hints but using the implicit size-hint methodology described in \S~\ref{sec:size_hints}. Most prompts follow the same theme of infilling either a \texttt{title} element or an element describing a headline (either through attributes or using the \texttt{meta} tag). For completeness, below is an example of a prompt that can do basic summarization.

\begin{table*}[htpb]
\centering
\small
\begin{tabular}{@{}lcccc@{}}
\toprule
Model        & Gigaword          & CNN/DM            & Reddit TIFU      & XSum             \\ \midrule
PEGASUS-0S   & 23.39/07.59/20.20  & 32.90/13.28/29.38 & 14.66/3.06/10.17 & 19.27/3.00/12.72 \\ 
\HTLM{}-Auto-NS & 27.56/10.17/24.57 & 33.40/13.45/30.10 & 06.71/1.98/07.86   & 15.15/2.54/10.91 \\
\HTLM{}-Auto-S  & 28.73/11.31/26.49 & 34.65/14.54/32.15 & 08.15/2.92/09.75   & 17.14/3.41/13.43 \\
\HTLM{}-Manual-S  & 31.61/10.80/28.60 & 38.51/16.10/33.89 & \textbf{15.81/2.98/10.54} & 22.34/4.12/14.56     \\ \midrule
\HTLiM{}-M-Manual  & 29.15/09.70/27.87 & 37.16/14.75/31.42 & 09.56/2.65/07.48  & 20.14/3.15/13.89    \\ 
\HTLiM{}-L-Manual  & \textbf{32.12/10.95/28.78} & \textbf{38.88/16.27/34.16} & 12.14/2.12/07.98 & \textbf{24.86/6.08/16.32}      \\ \bottomrule
\end{tabular}
\caption{\HTLiM{} results on zero-shot summarization. \HTLM{}-Manual denotes manually engineered prompts with size hints, while \HTLM{}-Auto-S and \HTLM{}-Auto-NS indicate auto-prompting with and without size hints, respectively. The metrics shown are ROUGE-1/ROUGE-2/ROUGE-L, respectively. \HTLiM{}-Large sets a new state-of-the-art on three news-based summarization datasets.}
\label{table:HTLM_summarization}
\end{table*}

We present our results in Table~\ref{table:HTLM_summarization}. Both \HTLiM{} models saw significantly less text than the \HTLM{} model, with 2.7TB of text. Furthermore, the prompts being used were tuned specifically for the \HTLM{} model and are being used with no changes for \HTLiM{}. With these challenges, we still see that \HTLiM{}-Large sets new state-of-the-art zero-shot summarization for three datasets. We attribute the performance degradation in Reddit-TIFU data to \HTLiM{} pre-training data only containing CC-NEWS and Wikipedia, which will not contain the type of summarizations needed for Reddit-TIFU.

\section{Fine-Tuning}
We next want to measure the quality of internal representations for the end goal of fine-tuning. We compare \HTLiM{} with a wide array of masked language model derived models such as T5 \citep{T5}, RoBERTa \citep{ROBERTA}, \HTLM{} \citep{HTLM1} tested on the standard GLUE benchmark \citep{GLUE}. For \HTLiM{} we look at three settings for fine-tuning; standard fine-tuning, better fine-tuning using adversarial methods \citep{RXF}, and better fine-tuning over prompts derived from \citet{HTLM1}. We delegate the specifics hyper-parameters of the fine-tuning experiments to \S~\ref{section:glue_hp}

\begin{table*}[h]
\centering
\small
\setlength{\tabcolsep}{3pt}
\begin{tabular}{@{}lcccccccr@{}}
\toprule
 &
  \begin{tabular}[c]{@{}l@{}}MNLI\\ Acc-m/mm\end{tabular} &
  \begin{tabular}[c]{@{}l@{}}QQP\\ Acc\end{tabular} &
  \begin{tabular}[c]{@{}l@{}}RTE\\ Acc\end{tabular} &
  \begin{tabular}[c]{@{}l@{}}QNLI\\ Acc\end{tabular} &
  \begin{tabular}[c]{@{}l@{}}MRPC\\ Acc\end{tabular} &
  \begin{tabular}[c]{@{}l@{}}CoLA\\ Mcc\end{tabular} &
  \begin{tabular}[c]{@{}l@{}}SST-2\\ Acc\end{tabular} & \quad \begin{tabular}[c]{@{}l@{}}\# Params\\ \quad\end{tabular}\\ \midrule

T5-Base & 87.1/86.2 & 89.4 & 80.1 &  93.7 & 87.5 & 51.1 &  95.2 & 220M\\
RoBERTA   & 90.2/-    & 92.2   & 86.6 & 94.7 & 89.1    & 68.0 & 96.4  & 330M  \\
RoBERTa-R3F        & 91.1/91.3         &  92.4         &  88.5    &     95.3  &   91.6        & 71.2      &  97.0    & 330M\\
BART-Large & 89.9/90.1 &  92.5 & 87.0 &  94.9 & 90.4 &  62.8 & 96.6 & 400M\\ 
\HTLM{}        & 90.3/91.4  &   92.6       &  87.1    &   95.1   & 90.8    &  64.3    &   96.9 & 400M \\  
\HTLM{}-R3F        & 91.4/92.1  &   92.8        &  89.1    &   95.4   & 91.5    &  69.4    &   97.1  & 400M \\
\HTLM{}-R3F-Prompt        & 91.6/91.2  &   92.9        &  89.4    &   95.7   & 91.7    &  69.8    &   97.3  & 400M\\ 
T5-Large & 89.9/89.6 & 89.9 & 87.2 & 94.8 & 89.9 & 61.2 &  96.3 & 770M \\
T5-3B &  91.4/91.2 & 89.7 & 91.1 & 96.3 & 90.0 & 67.1 &  97.4 & 3B \\
T5-11B & 92.2/91.9 & 90.6 & 92.8 & 96.9 & 90.4 & 71.6 &  97.5 & 11B \\ \midrule
\HTLiM{}-Medium &  89.9/89.7 & 89.6 & 89.1 & 93.1 & 86.5 & 63.1 &  94.9 & 2.7B \\
\HTLiM{}-Medium-Prompt &  90.8/91.0 & 89.9 & 90.5 & 95.1 & 89.9 & 66.2 &  96.3 & 2.7B \\
\HTLiM{}-Medium-RXF-Prompt &  90.9/91.1 & 90.0 & 90.7 & 95.3 & 90.0 & 67.1 &  96.9 & 2.7B \\ \midrule
\HTLiM{}-Large & 91.1/91.0 & 89.9 & 91.9 & 95.6 & 89.6 & 64.6 &  94.2 & 13B \\
\HTLiM{}-Large-Prompt & 91.5/91.4 & 90.1 & 92.4 & 96.2 & 90.1 & 70.9 &  97.1 & 13B \\
\HTLiM{}-Large-RXF-Prompt & 91.9/91.5 & 91.1 & 92.5 & 96.4 & 90.3 & 70.8 &  97.3 & 13B \\
\bottomrule

\end{tabular}
\caption{Results on the GLUE development set for various fine-tuning methods
applied to \HTLiM{}.}
\label{table:glue}
\end{table*}
We present our results in Table~\ref{table:glue}. Overall we see that both \HTLiM{} models are competitive against T5 given the same parameter setting. Furthermore, aligning with the results from \citet{HTLM1} we see that placing the natural language utterances of the various GLUE tasks into an HTML prompt while fine-tuning non-trivially improves end-finetuning performance. The following experiments show that the causally masked language modeling approach is not detrimental to learning fine-tunable representations, and neither is jointly modeling image tokens.

\comment{\section{Perplexity Analysis}\label{sec:ppl_analysis}}
\section{Ethical Considerations}
Prior work has explored the extent to which language models encode harmful gender and racial biases that parallel humans through the Word Embedding Association Test (WEAT) \citep{Caliskan2017SemanticsDA}, the Sentence Encoder Association Test (SEAT) \citep{May2019OnMS} and the Grounded-WEAT/Grounded-SEAT \citep{Ross2021MeasuringSB} metrics for multimodal language models \citep{Tan2019AssessingSA}.
Given the generative nature of \HTLiM{} in both the language and visual modalities, we used GWEAT/GSEAT to probe our model. Overall, we evaluated six bias tests for gender and seven bias tests for race and found that our family of \HTLiM{} models show significantly less bias than other models, speicifically VisualBERT \citep{visualbert} and ViLBert \citep{vilbert}.

\begin{table}[h]
\small\centering
\begin{tabular}{@{}llrrrr@{}}
\toprule
                              & Level & VisualBert & ViLBert & \HTLiM{}-Medium & \HTLiM{}-Large \\ \midrule
C6: M/W, Career/Family        &\texttt{S}    &\underline{1.05} &\underline{1.14}                       & 0.00              &  \underline{0.98}            \\
                              &\texttt{W}      &\underline{0.54} &\underline{0.51}         &   0.10            &       0.12       \\ \midrule
C8: Science/Arts, M/W         &\texttt{S}      &\underline{0.86} &\underline{1.05}                      &-0.09               & 0.42             \\
&\texttt{W}      &\underline{0.62} &\underline{0.14}                   &  0.08             &    0.07          \\ \midrule
C11: M/W, Pleasant/Unpleasant &\texttt{S}       &-0.74 &-0.84                    &0.00               & -0.64             \\
&\texttt{W}      &-0.66 &-0.31                    &   -0.20            &     -0.48         \\ \midrule
Double Bind: M/W, Competent   &\texttt{S}       &-0.10 &-0.04          &   0.01            &   -0.01           \\
&\texttt{W}      &-0.23 &\underline{0.30}                    &-0.07               &    -0.27          \\ \midrule
Double Bind: M/W, Likeable    &\texttt{S}       &-0.11 &-1.12            &-0.24               &    -0.59          \\
&\texttt{W}      &-0.60 &0.09                     &0.00               &    0.10          \\ \midrule
Occupations: M/W, Occupation  &\texttt{S}       &\underline{0.98} &\underline{1.82}                    &0.03               & \underline{0.62}             \\
&\texttt{W}      & \underline{0.91} & \underline{1.80}                  &0.00               & 0.58             \\
\midrule\midrule
Total Significant Bias Count & - & 5 & 6 & 0 & 2 \\ \bottomrule
\end{tabular}
\caption{Following \citet{Ross2021MeasuringSB} we present the results for all gender bias classes on answering the question: ``Do joint embeddings contain biases"? The numbers in our table represent effect sizes, and are underlined if their respective p-values are below 0.05. Each bias
type and model are tested three times against Word embeddings (W) and Sentence embeddings (S). }\label{table:ethics_gender}
\end{table}

\begin{table}[h]
\small\centering
\begin{tabular}{@{}llrrrr@{}}
\toprule
& Level & VisualBert & ViLBert & \HTLiM{}-Medium & \HTLiM{}-Large \\ \midrule
C3: EA/AA, Pleasant/Unpleasant & \texttt{W} & \underline{ 0.23} & \underline{ 0.14} & -0.44       & 0.10        \\
& \texttt{S}                                & \underline{ 0.31} & -0.14      & -0.057      & 0.05        \\ \midrule
C12: EA/AA, Career/Family & \texttt{W}      & -0.29      & \underline{ 0.43} & 0.117       & 0..23       \\
& \texttt{S}                                & -0.54      & \underline{ 0.34} & -0.049      & \underline{ 0.28}  \\\midrule
C13: EA/AA, Science/Arts &\texttt{W}       & 0.04       & \underline{ 0.21} & 0.325       & 0.12        \\
&\texttt{S}                                & \underline{ 0.12} & \underline{ 0.68} & \underline{ 0.169} & \underline{ 0.465} \\\midrule
Double Bind: EA/AA, Competent &\texttt{W}  & \underline{ 0.61} & \underline{ 0.87} & -0.535      & \underline{ 0.42}  \\
&\texttt{S}                                & 0.24       & 0.25       & 0.0         & 0.18        \\\midrule
Double Bind: EA/AA, Likeable &\texttt{W}   & 0.21       & -0.23      & -0.535      & 0.19        \\
&\texttt{S}                                & \underline{ 0.27} & -0.74      & -0.535      & 0.21        \\\midrule
Occupations: EA/AA, Occupation &\texttt{W} & -0.40      & 0.02       & -0.51       & 0.01        \\
&\texttt{S}                                & -0.41      & \underline{ 0.46} & -0.17       & 0.38        \\\midrule
Angry Black Woman Stereotype &\texttt{W}   & -0.07      & 0.26       & -1.89       & 0.21        \\
&\texttt{S}                                & -0.50      & \underline{ 0.47} & 0.0         & -0.10       \\ \midrule\midrule
Total Significant Bias Count & - & 4 & 5 & 1 & 3 \\ \bottomrule
\end{tabular}
\caption{Following \citet{Ross2021MeasuringSB} we present the results for all racial bias classes on answering the question: ``Do joint embeddings contain biases"? Our table uses the same annotations as Table~\ref{table:ethics_gender}. }\label{table:ethics_race}
\end{table}

We present our empirical results for gender and race bias in Table~\ref{table:ethics_gender} and Table~\ref{table:ethics_race} respectively. Overall, both \HTLiM{} have significantly less bias than other competing models, most likely due to our choice to use only Wikipedia and CC-NEWS articles as training sources (and recent CC-NEWS articles at that). Furthermore, we believe the fact that \HTLiM{}-Medium shows no to very little signs of bias can be an indicator of under-fitting as the large model is, unfortunately, able to show some bias from our training data.

We also qualitatively experiment with whether \HTLiM{} can be prompted to produce harmful or objectionable images. In general, we noticed it was incredibly hard to produce such content, additionally the lack of the ability to generate distinctive features of VQVAE-GAN acts to our benefit in terms of preserving privacy.

\section{Related Work}
Fundamentally our work is an extension of the \HTLM{} work proposed by \citet{HTLM1} to using the newly proposed causally masked objective, integrating images through VQVAE-GAN tokens, and scaling up over an order of magnitude. From there, the individual capabilities of our models are comparable to individual approaches.

For example, the conditional and unconditional image generation capabilities of our model are most similar in approach to DALL-E, which trains a left-to-right causal model over the concatenation of textual tokens and VQ-VAE visual tokens \citep{DALLE}. At the same time, the use of auto-regressive modeling in entity linking and disambiguation was proposed by the GENRE in \citet{GENRE}.

The method of tokenizing non-discrete modalities to use standard sequence modeling approaches have been extensively explored with DALL-E for images, Jukebox for Music \citep{jukebox} and vq-wav2vec for Speech \citep{vq-wav2vec}.

\section{Conclusion}
In this paper, we present the \HTLiM{} model, a causally masked trained language model that is capable of non-trivial zero-shot performance on a wide range of zero-shot uni- and cross-modal tasks. We first describe a new sequence modeling objective we call causally masked, enabling both full generative modeling with bidirectional context.

Through extensive experimentation, we show that as a single model \HTLiM{} can be prompted to recover the functionality of many other models being able to do image generation, image captioning, unconditional image generation, and more. Empirically we improve over state-of-the-art zero-shot summarization, entity linking, entity disambiguation, highlighting the structure from the hypertext during training. We show that representations learned by \HTLiM{} are not only useful for zero-shot prompting but for fine-tuning by fine-tuning \HTLiM{} and state-of-the-art for entity linking and entity disambiguation in general, all while staying highly competitive with T5 models on the GLUE benchmark. 

\bibliography{iclr2022_conference}
\bibliographystyle{iclr2022_conference}
\clearpage
\appendix
\section{Appendix}
\subsection{Model Architecture}\label{ssec:arch}
For model architecture we use the same exact architecture for \HTLiM{}-Medium and \HTLiM{}-Large as the dense 2.7B and 13B models described in \citet{efficient_lm_xlm_g}.

\begin{table}[h]
\centering
\begin{tabular}{@{}lrr@{}}
\toprule
                                   & \HTLiM{}-Large & \HTLiM{}-Medium \\ \midrule
--decoder-embed-dim                & 5120                            & 2560                             \\
--decoder-output-dim               & 5120                            & 2560                             \\
--decoder-input-dim                & 5120                            & 2560                             \\
--decoder-ffn-embed-dim            & 20480                           & 10240                            \\
--decoder-layers                   & 40                              & 32                               \\
--decoder-normalize-before         & True                            & True                             \\
--decoder-attention-heads          & 40                              & 32                               \\
--share-decoder-input-output-embed & True                            & True                             \\
--decoder-learned-pos              & False                           & False                            \\ \bottomrule
\end{tabular}
\caption{FairSeq architecture designation for \HTLiM{} models}
\label{table:architecture}
\end{table}
\subsection{Uniformity of VQVAE-GAN Tokens}\label{ssec:uniformity_tokens}
We plot a histogram of all image tokens in a subset of our data spanning 100k tokens. We see a somewhat clear uniformity in tokens used.

\begin{figure}[h]
    \centering
    \includegraphics[width = 5.4in]{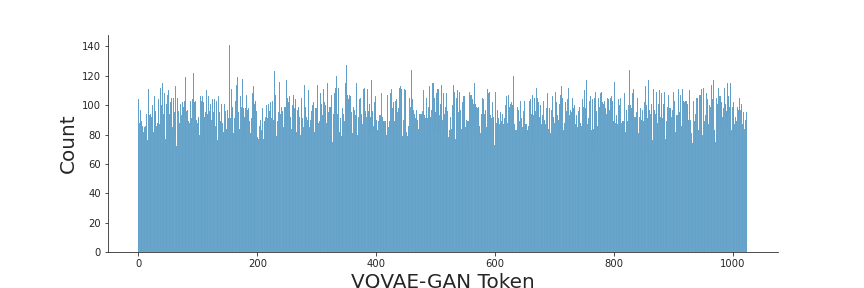}
    \caption{Histogram of VQ-VAE-GAN Tokens in the \HTLiM{} Training Dataset.}
    \label{fig:uniformity_tokens}
\end{figure}

\subsection{Finetuning GLUE Hyper-Parameters}\label{section:glue_hp}
% commenting out tables due to compile time
For our fine-tuning GLUE related experiments with the RXF method we use the following hyper-parameters.

\begin{table}[h]
\centering
\begin{tabular}{@{}llllllll@{}}
\toprule
Hyper Parameter & MNLI   & QNLI  & QQP    & SST-2 & RTE  & MRPC & CoLA \\ \midrule
Learning Rate   & 5e-6   & 5e-6  & 5e-6   & 5e-6  & 1e-5 & 1e-5 & 1e-5 \\
Max Updates     & 123873 & 33112 & 113272 & 20935 & 3120 & 2296 & 5336 \\
Max Sentences   & 8      & 8     & 32     & 32    & 8    & 16   & 16   \\
\bottomrule  
\end{tabular}
\caption{Task specific hyper parameters for GLUE experiments}
\end{table}

\begin{table}[h]
\centering
\begin{tabular}{@{}ll@{}}
\toprule
Hyper parameter & Value              \\ \midrule
Optimizer       & Adam               \\
Adam-betas      & (0.9, 0.98)        \\
Adam-eps        & 1e-6               \\
LR Scheduler    & polynomial decay   \\
Dropout         & 0.1                \\
Weight Decay    & 0.01               \\
Warmup Updates  & 0.06 * max updates \\
\bottomrule
\end{tabular}
\quad
\begin{tabular}{@{}ll@{}}
\toprule
Hyper parameter & Value           \\ \midrule
$\lambda$          & [0.1, 0.5, 1.0, 5.0] \\
Noise Types     & [$\mathcal{U}$, $\mathcal{N}$] \\
$\sigma$        & $1e-5$\\ 
\bottomrule
\end{tabular}
\caption{Hyper parameters for fine-tuning experiments on GLUE}
\end{table}

\end{document}

%% file: math_commands.tex
\usepackage{amsmath,amsfonts,bm}

\def\eqref#1{equation~\ref{#1}}
\def\1{\bm{1}}

\DeclareMathAlphabet{\mathsfit}{\encodingdefault}{\sfdefault}{m}{sl}
\SetMathAlphabet{\mathsfit}{bold}{\encodingdefault}{\sfdefault}{bx}{n}